\documentclass{article} % For LaTeX2e
\usepackage{iclr2023_conference,times}

\usepackage{amsmath,amsfonts,bm}

\def\eqref#1{equation~\ref{#1}}
\def\1{\bm{1}}

\DeclareMathAlphabet{\mathsfit}{\encodingdefault}{\sfdefault}{m}{sl}
\SetMathAlphabet{\mathsfit}{bold}{\encodingdefault}{\sfdefault}{bx}{n}

\pdfoutput=1

\usepackage{hyperref}
\usepackage{url}
\usepackage{tikz}
\usepackage{xcolor}
\usepackage{tcolorbox}
\usepackage{color,xcolor}

\usepackage{amssymb}
\usepackage{amsopn}
\usepackage{graphicx}
\usepackage{multirow}
\usepackage[english]{babel}
\usepackage{bm}
\usepackage{array}
\usepackage{setspace}
\usepackage{todonotes}

\usepackage{xspace}
\usepackage{amsfonts}
\usepackage{array,multirow}
\usepackage{pgfplots}
\usepackage{tikz}
\usepackage{verbatim}
\usepackage{subfig}
\usepackage{arydshln}
\usepackage{booktabs}
\usepackage{fancyhdr}

\title{RecycleGPT: An Autoregressive Language Model with Recyclable Module}

\author{{\bf Yufan Jiang\thanks{Equal contribution, correspondence to 
\{jiangyufan2018,qiaozhihe2022\}@outlook.com}  , \ Qiaozhi He$^{*}$, \ Xiaomin Zhuang, \ Zhihua Wu,} \\ {\bf Kunpeng Wang$^{1}$, \ Wenlai Zhao$^{1}$, \ Guangwen Yang$^{1}$} \\
$^{1}$Department of Computer Science and Technology, Tsinghua University, Beijing, China \\
}

\iclrfinalcopy % Uncomment for camera-ready version, but NOT for submission.
\begin{document}

\maketitle

\begin{abstract}
% The abstract paragraph should be indented 1/2~inch (3~picas) on both left and
% right-hand margins. Use 10~point type, with a vertical spacing of 11~points.
% The word \textsc{Abstract} must be centered, in small caps, and in point size 12. Two
% line spaces precede the abstract. The abstract must be limited to one
% paragraph.
Existing large language models have to run $K$ times to generate a sequence of $K$ tokens. 
In this paper, we present RecycleGPT, a generative language model with fast decoding speed by recycling pre-generated model states without running the whole model in multiple steps.
Our approach relies on the observation that adjacent tokens in a sequence usually have strong correlations and the next token in a sequence can be reasonably guessed or inferred based on the preceding ones.
Experiments and analysis demonstrate the effectiveness of our approach in lowering inference latency, achieving up to 1.4x speedup while preserving high performance.
% Trained models, analysis code, training code, and training data can be found at ..
\end{abstract}

\section{Introduction}
Large language models (LLMs) \citep{brown2020language,openai2023gpt4,touvron2023llama,chowdhery2022palm,biderman2023pythia,smith2022using} have revolutionized the field of natural language generation for their abilities in generating satisfactory text across various application domains. The excellent performance benefits greatly from the scaling of model size (100B+ parameters), but at the same time, the fact remains that a single decoding step gets slower as the model gets larger.

In addition to the immense computation introduced by larger models, a larger memory footprint is also a major factor causing slower inference of LLMs  \citep{dao2022flashattention,pope2023efficiently}.
% First, large language models have a large memory footprint due to both the trained model parameters and the temporary state needed during decoding. 
% Furthermore, the KV cache is also stored in memory for the duration of decoding.
This large memory footprint includes the trained model parameters, the temporary state used during inference, and in addition to these, the KV cache is also stored in memory.
% Secondly, the large memory footprint leads to considerable memory traffic to load the parameters and KV cache from high-bandwidth memory (HBM) into the compute cores at each step, requiring high total memory bandwidth to meet a given latency target. 
At each decoding step, it has to load the parameters and KV cache from high-bandwidth memory (HBM) into the compute cores which results in significant memory traffic and thus, high total memory bandwidth is required to meet a given latency target.
In other words, the speed of generating tokens in LLMs is primarily limited by how fast it can access memory \citep{shazeer2019fast,pope2023efficiently,chen2023accelerating}. 
And the time to generate each token is roughly proportional to the number of model parameters. 
% Serving larger language models also need multiple devices to work together in parallel, which adds communication overhead and increases the amount of computing resources required. 
Since each new token generated by the model depends on the previous tokens, many calls to the transformer model are necessary to generate an entire sequence.

% Generally, LLMs adopt auto-regressive decoding. At test time, the system produces one word each time until an end symbol is reached, which means decoding $N$ tokens needs to run the model $N$ times. 
% Generally, at test time, LLMs produce one word each step until an end symbol is reached, which is also called auto-regressive decoding.
% This serially approach preserves left-to-right dependencies between tokens for better generated results, but leads to sub-optimal decoding speed and limited GPU utilization.
 
To make inference more efficient, several works are proposed. 
% The core idea of these works is how to shorten the computation path of the model or reduce the number of model calls.
The core idea of these works is how to reduce the memory footprint and alleviate memory traffic problems.
For example, distillation \citep{hinton2015distilling}, sparcification \citep{jaszczur2021sparse}, quantization \citep{shen2020q,zafrir2019q8bert} and sharing weights \citep{xiao2019sharing,zeng2021recurrent} are proposed to reduce the model size. 
Adaptive computation \citep{sukhbaatar2019adaptive,schwartz2020right} aims to use fewer computing resources for easier inference steps.
Multi-Query Attention \citep{shazeer2019fast,ainslie2023gqa} shares the keys and the values to reduce the size memory bandwidth requirements while Flash Attention \citep{dao2022flashattention} uses a small amount of computation to reduce the number of memory reads/writes.
Though the above works propose effective approaches, they usually require changing the model architecture or attention algorithm, adding more training tasks, and re-training these complicated models.
Recently speculative decoding methods have become popular \citep{leviathan2023fast,chen2023accelerating,miao2023specinfer}.
To reduce the number of executions of the large model, they employ a two-step approach: first, an efficient small model speculatively generates the simpler parts of the text; then, a large model is used to validate those parts, rather than having the large model generate the entire text alone.
% At first, a short draft tokens are generated by parallel decoding or a more efficient auto-regressive model. Then a larger model are employed to decide which subset of draft tokens are kept.
This idea is simple and convenient and also has been integrated to open-source frameworks.
However, the selection of efficient models is still an open question. Using the small version of LLMs may be one solution while it still needs sequence-level distillation.
 
% Unfortunately, a single decode step from these larger models is significantly
% slower than a step from their smaller counterparts, and making things worse, these steps are done serially - decoding K tokens takes K serial runs of the model.
% In this work, we observe that the inference bottleneck of large Transfomers..
% This motivate us to design a model predict 
% model predict continuously
% Another solution to this problem is to minimize the data transmission time. 
Naturally, adjacent tokens in a sequence have strong correlations.
That is to say, in many cases, the next token in a sequence can be reasonably guessed or inferred based on the preceding ones.
This phenomenon leads us to investigate an efficient decoding method in another research direction, with the goal of generating as many tokens as possible under the same amount of memory processing budget.
% This experience motivate us to present an new architecture to accelerate language model decoding.....
We propose {\bf RecycleGPT}, a novel language model architecture that is inherently capable of fast decoding by recycling pre-generated model states.
% We propose a novel language model architecture that is inherently capable of fast decoding by generating multiple tokens in each step.
% We propose an new architecture to accelerate language model decoding by generating multiple tokens in each step.??
In our approach, we modify the original language model by adding an additional recyclable module that predicts the next several tokens using previously generated states without running the entire model multiple times, which can also be viewed as a recycling process.
% The recycling module is made up of a stack of transformer-based layers and we also improve the attention mechanism for achieving more efficient representations to make prediction.
The recyclable module is made up of a stack of transformer-based layers for achieving more efficient representations to make predictions.
During inference, this module can be used with the standard language model decoding pipeline in various ways.
In this paper, we choose to use them alternately (i.e., generating every two tokens requires running the complete model once) and leave exploring more strategies for future work.
Despite its simple architecture, the recyclable module can effectively represent contextual information and make accurate predictions, thereby achieving the goal of accelerating the decoding process.
% Our technique can be applied to existing pre-trained models through a straightforward fine-tuning process, and the speed-up rate can be adjusted by a skip parameters.

We evaluate the RecycleGPT on a set of standard benchmarks.
% show an average of 1.3X speed-up
It achieves a 1.4x speedup over the standard language model, yet with no loss in performance.
More importantly, it is orthogonal to previous methods and is straightforwardly applicable to different LLMs.
% leads to a 2–2.5× speedup
The main contributions of this work are summarized as follows:
\begin{itemize}
    \item We propose a novel generative language model RecycleGPT and release RecycleGPT-1.3B. Compared to standard language models, our model achieves 1.4x speedup with only 15\% extra parameters introduced, while maintaining comparable performance on downstream tasks. In the future, we will release variants of RecycleGPT in different sizes.
    \item Our recycling method is flexible and scalable, which can be applied to different pre-trained models. Moreover, the size of the recyclable modules and the generation strategy can be adjusted to achieve the desired speedup performance.
\end{itemize}

\section{Background}
In this section, we provide some background on the memory cost at inference time.
We also give a brief introduction to the auto-regressive language model. 
%Fig1 shows the overall framework.
% XXX is composed of 
\subsection{Inference Memory Cost}
As the model scale continues to explode exponentially, language model decoding becomes highly costly and inefficient.
Except that larger models introduce more tensor computations that take up a certain amount of time, the memory transfer also occupies a significant portion of time.
Generally, large language models have a large memory footprint for storing both model parameters and KV cache which are usually stored in on-device high-bandwidth memory (HBM).
These tensors need to be transferred from HBM to the compute cores each forward pass which takes a certain amount of time.
And since the auto-regressive language models generate one token each step until the end symbol is reached, many calls to the language model are necessary to
generate an entire sequence.
According to \citet{pope2023efficiently}, at small batch sizes and sequence lengths, loading weights takes the most time, while loading the KV cache dominates inference time at a large scale.
Moreover, larger language models need multiple devices to work together in parallel, which also adds communication overhead.
Thus, how to reduce the memory size and transfer frequency is another key factor in accelerating the model decoding process.

% how to reduce this part
\subsection{Auto-regressive language model}
\begin{figure}[!t]
\centering
\includegraphics[width=1.0\linewidth]{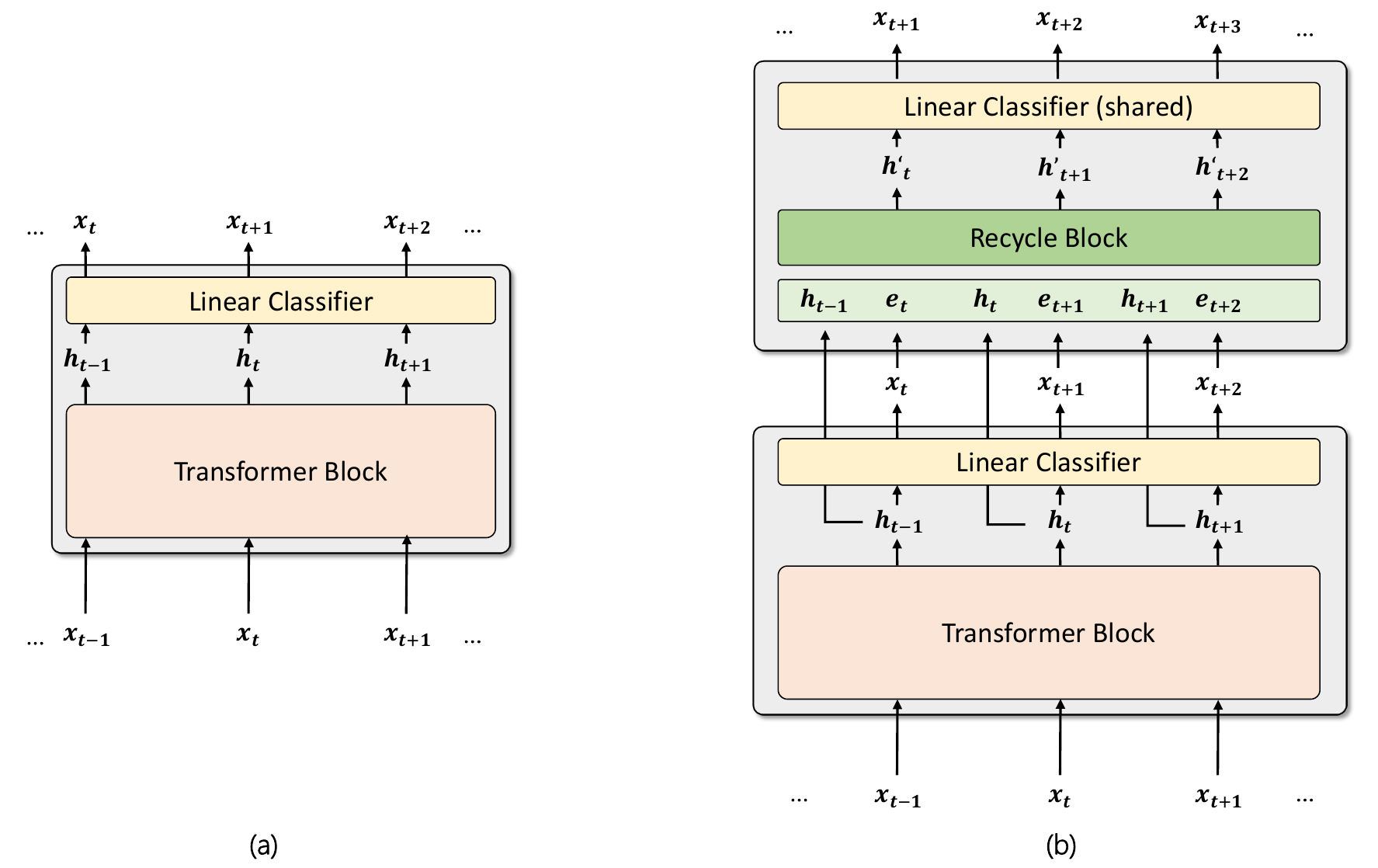}
\caption{
{\bf Model architecture of standard GPT and RecycleGPT}.
{
}
}
\label{fig_model}
\end{figure}

% Given an initial prompt sequence $X = \{x_1,..., x_n\}$, an auto-regressive language model (Figure \ref{fig_model} (a)) factors the distribution over possible future sequence $\{x_{n+1},..., x_{t+1}\}$ into a chain of conditional probabilities with a left to right causal structure:
% \begin{equation}
%    \mathcal{P_{AR}}(X_{n+1:t+1}|X_{1:n};\theta) = \prod^{t+1}_{i=n+1}  p(x_i|x_{1:i-1};\theta_{AR}),
% \end{equation}
Given a corpus of tokens $X = \{x_1,..., x_n\}$, an auto-regressive language model (Figure \ref{fig_model} (a)) factors the joint probability into a chain of conditional probabilities with a left to right causal structure:
\begin{equation}
   \mathcal{P_{AR}}(X;\theta_{AR}) = \prod^{n}_{i=1}  p(x_i|x_{<i};\theta_{AR}),
\end{equation}
For most LLMs, transformer-based models are used to capture the above causal structure of the output distribution. 
Generally, in transformer, there are $L$ identical stacked layers. Each of them is composed of a self-attention sub-layer and a feed-forward sub-layer (FFN). Both of them are equipped with a residual connection and a layer normalization unit. For more details, we refer the reader to \citet{NIPS2017_3f5ee243}.
When generating the token $x_{t+1}$, a distribution over vocabulary tokens is computed via a softmax-normalized linear classifier $\textbf{W}_L$ with $h_{t}^{L}$ as input:

\begin{equation}
   p(x_{t+1}|h_{t}^{L}) = {\rm softmax}(\textbf{W}_L h_{t}^{L}),
   \label{dis1}
\end{equation}
where $h_{t}^{L}$ is the decoder state of the last layer of the transformer model.
Finally, the (greedily chosen) prediction $x_{t+1}$ can be written as:

\begin{equation}
   x_{t+1} = {\rm argmax}\ p(x_{t+1} | h_{t}^{L})
   \label{dis2}
\end{equation}

At the same time, maximum likelihood training with a cross-entropy loss can be applied at each decoding step:

% \begin{equation}
%    \mathcal{L}_1 = {\rm log} \mathcal{P_{AR}}(X_{n+1:t+1}|X_{1:n};\theta_{AR}) = \sum^{t+1}_{i=n+1}  {\rm log}\ p(x_i|x_{1:i-1};\theta_{AR}),
% \end{equation}

\begin{equation}
   \mathcal{L}_1 = {\rm log} \mathcal{P_{AR}}(X;\theta_{AR}) = \sum^{n}_{i=1}  {\rm log}\ p(x_i|x_{<i};\theta_{AR}),
\end{equation}

Though the transformer structure shows strong generative capabilities and high parallelism during training.
It has been pointed out that the auto-regressive format is highly memory bandwidth bound and is difficult to leverage modern accelerator hardware effectively \citep{chen2023accelerating,shazeer2019fast}.
This kind of memory-bound model generates one word per call, hence generating multiple words in sequence induces high latency and it gets worse as the number of model parameters increases.

\section{RecycleGPT}
% Based on the observation that 
In order to minimize the time spent on both memory transfer and computation, we aim to reduce the number of calls of the full-parameter language model.
Instead of always making predictions according to the previous token, we propose a simple but effective solution.
Based on the assumption that neighboring tokens are highly correlated and interdependent, we directly recycle the representation of the current token to predict the following $m$ consecutive tokens without feeding each predicted token into the language model step by step. 
In this work, we only focus on the case where $m$ is 2 and we leave exploring this for future work.
% Though the .... due to the self-attention mechanism, in our preliminary experiments, it 

Thus we introduce RecycleGPT, a new generative language model.
Figure \ref{fig_model} shows the overall framework.
RecycleGPT includes a simple but effective recyclable module that is made up of a stack of $N$ identical transformer layers.
We use these few layers to predict the next token directly without feeding the current hidden state to the bottom of the language model and run the whole model to make predictions.
The design of these layers should consider how to strengthen the dependencies between discontinuous tokens, i.e. two tokens with one space and we will give a detailed introduction in the next section.
%Taking the generation of the token $x_{t+2}$ as an example, 
% When generating the token $x_{t+1}$, XXX takes the decoder state $h_{t-1}^{L}$ of token $x_{t-1}$ and embedding $e_{t}$ of token $x_{t}$ as input.
When generating token $x_{t+1}$, decoder state $h_{t-1}^{L}$ and embedding $e_{t}$ of token $x_{t}$ are passed through the recyclable module (Recycle) to obtain alternation state $h'_{t}$ which can be fed into linear classifier layer to predict $x_{t+1}$ like Eq.(\ref{dis1}) and Eq.(\ref{dis2}):

\begin{equation}
\begin{aligned}
x_{t+1} &= {\rm argmax}\ p(x_{t+1}|h'_{t}), \\
p(x_{t+1}|h'_{t}) &= {\rm softmax}(\textbf{W}_L h'_{t}), \\
h'_{t} &= {\rm Recycle}(g(h_{t-1}^{L},e_{t})),
\label{l2}
\end{aligned}
\end{equation}
% \begin{align}
%     p(x_{t+1}|h_{t}^{xxx}) = {\rm softmax}(\textbf{W}_L h_{t}^{xxx}), \\
%     h_{t}^{xxx} = {\rm xxxblock}(g(h_{t-1}^{L},e_{t})).
% \end{align}

where $g(\cdot,\cdot)$ is the function to integrate two streams of representations. 
We adopt the concatenating method for combining these two representations which is also introduced in the next section. 
According to \ref{l2},  we use the following objective to optimize the parameters of Recycle:

% \begin{equation}
%    %\mathcal{L}_2(U) = \sum_{i} {\rm log} p(u_i|u_{1:i-2},e_i;\theta_1, \theta_2),
%     \mathcal{L}_2 = {\rm log} \mathcal{P}_{Recycle}(X_{n+1:t+1}|X_{1:n};\theta_{Recycle}) = \sum^{t+1}_{i=n+1}  {\rm log}\ p(x_i|x_{1:i-2};\theta_{Recycle}),
% \end{equation}

\begin{equation}
   %\mathcal{L}_2(U) = \sum_{i} {\rm log} p(u_i|u_{1:i-2},e_i;\theta_1, \theta_2),
    \mathcal{L}_2 = {\rm log} \mathcal{P}_{Recycle}(X;\theta_{Recycle}) = \sum^{n}_{i=1}  {\rm log}\ p(x_i|x_{<i-1};\theta_{Recycle}),
\end{equation}

% From xxx, we get a fast version hidden state $h_{t}^{xxx}$ which can be fed into Eq.(\ref{dis}) directly:
In this work, we build RecycleGPT, a transformer based language model with a recyclable module, and train it from scratch.
% In this work, we build recyclable module on a transformer based language model and train them both from scratch. 
Thus, the training objective of our language model can be formulated as:

\begin{equation}
   \mathcal{L}_3(X) = \mathcal{L}_1(X) + \lambda * \mathcal{L}_2(X),
\end{equation}

Where $\lambda$ is a hyper-parameter to balance the effect of each loss term. 
%Our architecture of XXX also follows LLaMA's.
% while each generated token to the language model step by step.
% And during inference, we also need to select which 
% have two different strategies to select which 

\begin{figure}[!t]
\centering
\includegraphics[width=1.0\linewidth]{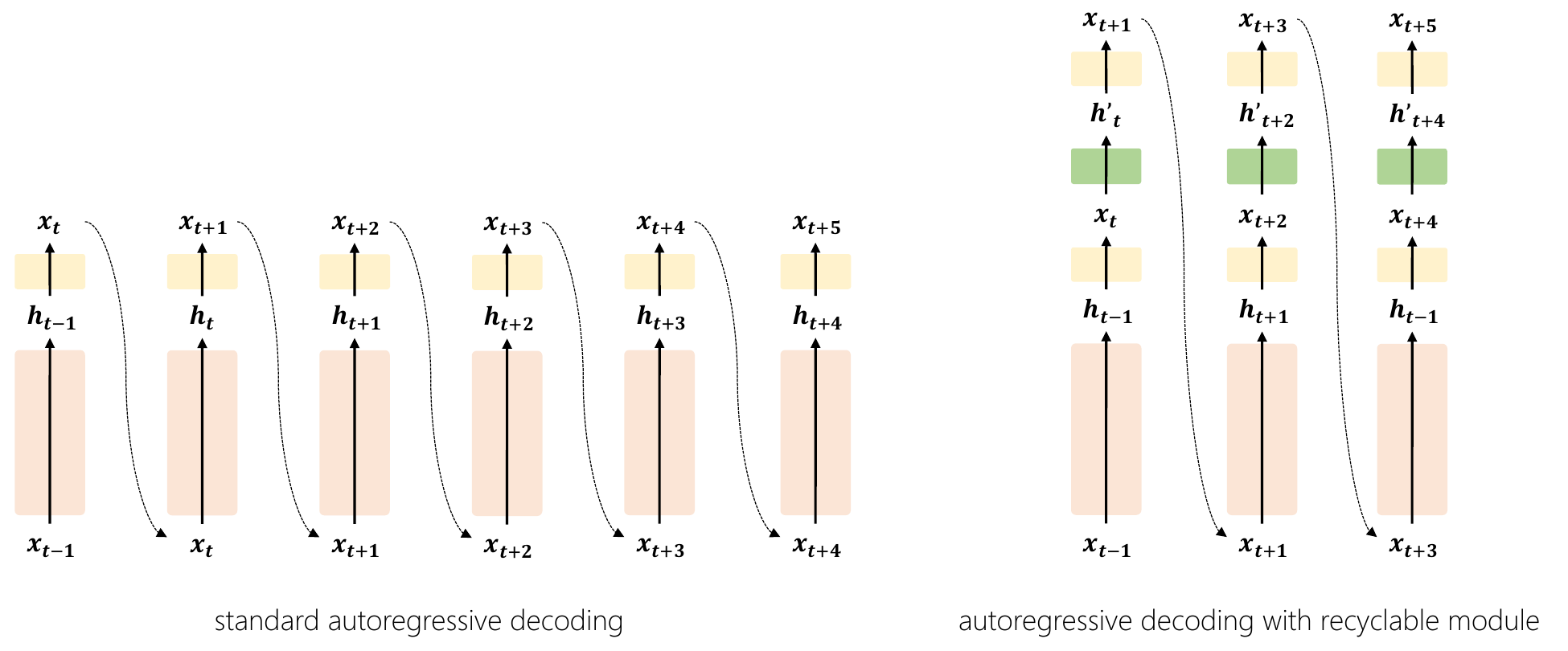}
\caption{
{\bf Illustration of the difference between standard autoregressive decoding and autoregressive decoding using a recyclable module.}.
{
The orange block indicates one forward call of the whole language model while the green one indicates the call of the recyclable module. The amount of computation and memory footprint required by the green part is far less than that of the orange part. When using an alternating decoding strategy, we see that the recyclable module can save a significant amount of time. The yellow block indicates the final output classifier.
}
}
\label{fig_decoding}
\end{figure}

For easier understanding, we illustrate the difference between auto-regressive decoding and our methods in Figure \ref{fig_decoding}.
Rather than generating $h_{t}^{L}$ through the complete execution of the language model using token $x_{t}$ as the input.
We generate $h'_{t}$ by the recyclable module with the hidden state of the last step and the token it predicted.
After obtaining $h'_{t}$, we can directly use it to predict token $x_{t+1}$.
% Instead of producing $h_{t}^{L}$ by running the whole language model with token $x_{t}$ as input, we generate it by the xxx module with the hidden state of last step and the  
Recycle module can speed up decoding due to its compact structure compared with whole language model layers.
Based on the $m$ being set to 2 in this work, we adopt a simple strategy of alternately using $h'_{t}$ and $h_{t}^{L}$ to generate the next word for each inference step.
Moreover, Our RecycleGPT can also perform standard auto-regressive decoding without using the recyclable module which we denote as RecycleGPT-std in the experiments section.
% for fast and simple
% skips the computation at 
% For easier understanding, we illustrate the 
% We leave exploring this for future work
% Here, by sweeping over the input text, we get short chunks (n-grams) and use their representations to approximate the
% hidden states of BERT’s lower layers.
% \subsection{Skip-Decoding Attention}

\subsection{Recyclable Module}
In this section, we give a detailed description of the Recyclable module.
This module is introduced to generate the substitute for the original decoder state $h^L_{t}$ which can be used to predict the next token.
The recyclable module helps the language model exploit the dependencies between discontinuous words.
% The XXX is designed to make a language model predicting multiple tokens with a single-step decoder state.
There are various ways to construct this module such as GRU \citep{cho2014properties}, LSTM \citep{graves2012long}, and FFN \citep{NIPS2017_3f5ee243}.
In this paper, we employ a number of transformer layers \citep{NIPS2017_3f5ee243} for better capturing the relationship between discontinuous tokens.
Recently, there are many variants of transformer layer, and we choose LLaMA, \citep{touvron2023llama} a stronger one among them.
It leverages various improvements that are subsequently proposed,
or used in different large language models, like RMSNorm \citep{zhang2019root}, swiGLU activation function \citep{shazeer2020glu} and rotary embeddings \citep{su2021roformer}.

Figure \ref{fig_model} depicts the structure of the recyclable module.
Before fed into the recyclable module, $h'_{t-1}$ and $e_t$ are concatenated along the length dimension at first. 
And we also set position embeddings for them.
Given the merged sequence $\{h_0, e_1, h_1, e_2...,h_{t}, e_{t+1}, h_{t+1}, e_{t+2}\}$, the corresponding position embedding is set to $\{0, 1, 1, 2...,t, t+1, t+1, t+2\}$ for both standard absolute position embeddings and rotary embeddings. 
% where $h_{t-1}$ and $e_t$ use the same one.
Then, the concatenation of two representations is passed through a stack of $N$ pre-norm LLaMA layers \citep{wang-etal-2019-learning-deep,touvron2023llama} which consist of self-attention sub-layers and feed-forward sub-layers to get the final representation of recyclable module.
The number of recyclable module layers $N$ in this work is adjustable based on hardware constraints to achieve the desired speedup performance.
For example, when $N$ is set to 6, the recyclable module introduces approximately 15\% extra parameters and achieved a 40\% decoding speedup when using the alternating decoding strategy.
Compared to other methods that reduce the number of model invocations, such as speculative decoding \citep{chen2023accelerating,leviathan2023fast}, our method is fine-grained while also being orthogonal to their methods, allowing further acceleration on top of them.

% \begin{figure}[!t]
% \centering
% \includegraphics[width=1.0\linewidth]{fig1_v4_crop.pdf}
% \caption{
% {\bf Model architecture of standard GPT and RecycleGPT}.
% {
% }
% }
% \label{fig_model}
% \end{figure}

% adjustable 
% It is obvious that .. 6 layers add small parameters ... speed up.
% 1)LLaMA structure
% 2)position embedding
% 3)self attention for better
% 4)It is obvious that fast than decoding from
% \subsection{Pre-training of XXX}
% % same structure with llama and optimized with 2 loss: 1) standard casual lm loss and skip loss
% % XXX can be trained with 

\section{Experiments}
\subsection{Experimental Setups}
\paragraph{Training Data.}
 Our model is trained on the Pile \citep{gao2020pile,biderman2022datasheet}, a carefully selected group of English language datasets for training large language models. The Pile is well-suited for training large autoregressive transformers. The reason we choose this public dataset is that it can achieve higher downstream performance than other popular datasets like C4 \citep{raffel2020exploring} and OSCAR \citep{suarez2019asynchronous}.
Additionally, this dataset has been widely utilized by state-of-the-art models including GPT-NeoX-20B \citep{black2022gpt}, Megatron-Turing NLG 530B \citep{smith2022using}, OPT \citep{zhang2022opt} and Pythia \citep{biderman2023pythia}.
We use the BPE tokenizer developed by \citet{touvron2023llama}. Overall, our entire training dataset contains 360B tokens after tokenization.

\paragraph{Training.}
We select LLaMA \citep{touvron2023llama} as our backbone and train a 1.3 billion parameter model.
The RecycleGPT has 24 layers with 2048 hidden units and 32 attention heads. 
We set $N=6$ for the recyclable module and it introduces 15\% parameters to the original model respectively.
$\lambda$ is set to 1 in this work.
Our model is trained using the Adam optimizer with the following hyper-parameters: $\beta_1 = 0.9$,  $\beta_2 = 0.95$.
Inspired by some of the latest research works \citep{biderman2023pythia,brown2020language}, we use a larger batch size than the standard language model. As a result, we use a batch size of 1280 samples, with each sample having a sequence length of 2048 tokens for our model.
The detail of the pre-training settings can be found in Appendix \ref{tab_settings_pretraining}.
When using RecycleGPT for decoding, we can choose to use the recyclable module for alternating generation denoted as RecycleGPT-rec, or perform standard auto-regressive decoding denoted as RecycleGPT-std.

We adopt several efficient implementations to improve training speed. 
First, we use flash attention \citep{dao2022flashattention} during training to increase device throughput.
In addition, we leverage the Zero Redundancy optimizer (ZERO) \citep{rajbhandari2020zero} to efficiently scale across multi-machine.
We also use data parallelism \citep{goyal2017accurate} and tensor parallelism \citep{shoeybi2019megatron} to optimize performance.
% We train our models using the open-source library xx developed by TecoAI. The model is trained on xxx SWAI GPU...

\paragraph{Evaluation.}
We empirically evaluate RecycleGPT on several common language modeling benchmarks in both zero-shot and few-shot settings. 
% six standard common sense reasoning benchmarks, namely PIQA \citep{bisk2020piqa}, WinoGrande \citep{sakaguchi2021winogrande}, ARC easy and challenge \citep{clark2018think}, SciQ \citep{welbl2017crowdsourcing}, LogiQA \citep{liu2020logiqa} and Lambada \cite{storks2019recent} in both zero-shot and few-shot settings.
\begin{itemize}
    \item {\bf Zero-Shot}. we provide the model with a textual description of the task and a test example as context. The model is then tasked with either generating an open-ended answer or ranking a set of multiple-choice answers.
    \item {\bf Few-Shot}. we provide the model with a few examples of the task and a test example as context. The model is then tasked with either generating an open-ended answer or ranking a set of multiple-choice answers.
\end{itemize}

% The above datasets include Cloze and Winograd style tasks, as well as multiple choice question answering and 
We use the Language Model Evaluation Harness \citep{gao2021framework} to run evaluations and use the same evaluation metric with \citet{biderman2023pythia} for a fair comparison.
Our efficiency metric is the speedup of the whole model for generating the full sequence with different lengths.
We perform decoding on a single A100 GPU with 200 examples and the results come from the average of 3 individual runs.
When decoding we use the greedy search method.

\paragraph{Baselines.} For a fair comparison, we collected existing open-source language models with around 1.3B parameters as baselines that are listed below:
1) OPT \citep{zhang2022opt}, a suite of decoder-only pre-trained transformers ranging from 125M to 175B parameters, and the architecture, tokenizer is almost identical to the standard GPT model. 2) Pythia \citep{biderman2023pythia} a suite of LLMs all trained on Pile datasets ranging in size from 70M to 12B parameters. Pythia improve the original architecture with a few notable deviations based on recent advances in best practices for large-scale language models. Since the LLaMA \citep{touvron2023llama} did not release a 1.3B parameter baseline, we revisit a llama-1.3B ourselves using the pile dataset.

\subsection{Results}
\paragraph{Common Sense Reasoning.}
We evaluate our models on standard common sense reasoning benchmarks, namely PIQA \citep{bisk2020piqa}, WinoGrande \citep{sakaguchi2021winogrande}, ARC easy and challenge \citep{clark2018think}, SciQ \citep{welbl2017crowdsourcing}, LogiQA \citep{liu2020logiqa} and Lambada \cite{storks2019recent} in the zero-shot setting.

\begin{table*}[!ht]
\centering
\small
\begin{spacing}{1.1}
\setlength{\tabcolsep}{1.3mm}{
\begin{tabular}{lccccccccc}
\toprule
{\bf Model} & {} & {\bf PIQA} & {\bf ARC-c} & {\bf ARC-e} & {\bf WinoGrande} & {\bf Lambada} & {\bf SciQ} & {\bf LogiQA} & {\bf Avg} \\
\midrule
OPT \dag & 1.3B & 71.7 & 23.7 & 57 & 59.7 & 57.9 & 84.5 & 22.3 & 53.8 \\
Pythia \dag & 1.4B & 70.5 & 25.3 & 59.4 & 56 & 59.2 & 87.3 & 22.4 & 54.3 \\
\midrule
OPT & 1.3B & 71.6 & 23.3 & 57.2 & 59.2 & 57.9 & 84.3 & 22.4 & 53.7 \\
Pythia & 1.4B & 70.8 & 26.0 & 60.6 & 57.3 & 61.7 & 86.6 & 21.2 & 54.9 \\
GPT-Neo & 2.7B & 72.2 & 27.6 & 61.1 & 58.0 & 62.2 & 89.2 & 19.7 & 55.7 \\
\midrule
LLaMA-ours  & 1.3B & 70.2 & 24.5 & 56.9 & 54.8 & 58.0 & 85.2  & 20.9 & 52.9 \\
RecycleGPT-std & 1.3B & 70.6 & 25.0 & 57.1 & 55.4 & 58.1 & 87.5 & 20.7 & 53.5 \\
RecycleGPT-rec & 1.5B & 68.7 & 24.6 & 56.7 & 55.3 & 57.6 & 86.4 & 23.8 & 53.3 \\
% RecycleGPT-std & 1.3B & 70.2 & 24.7 & 56.6 & 55.2 & 59.1 & 86.6 & 22.2 & 53.5 \\
% RecycleGPT-rec & 1.5B & 68.9 & 24.2 & 54.9 & 55.9 & 57.7 & 86.2 & 24.0 & 53.1 \\
\bottomrule 
\end{tabular}}
\end{spacing}
\caption{
\label{tab1}
{\bf Zero-shot performance on Common Sense Reasoning tasks.}
Models with {\dag} denote that we directly report the scores from the Pythia paper \cite{biderman2023pythia}, and others are from our implementation. Due to introducing the recyclable module, the number of parameters in our RecycleGPT has become 1.5B.
}
\end{table*}

In table \ref{tab1}, we report performance on six common sense reasoning benchmarks. 
On these benchmarks, our self-trained model and reproduced baseline model achieved competitive results with existing open-source models of the same size.
The performance gap on some benchmarks may be caused by the differences in training data and the tokenizer we used.
Compared to our own baseline, RecycleGPT using a standard decoding strategy (RecycleGPT-std) achieved comparable results, which proves that our recyclable module does not degrade the language model performance. 
Meanwhile, using the alternating decoding strategy (RecycleGPT-rec) can achieve 1.4x decoding acceleration with only less than one percentage point performance drop.
In actual use, the decoding strategy can be chosen based on acceleration requirements. 
We will also provide more combinations such as multiple decoding strategies and different recyclable module sizes for selection in the future.
% First, RecycleGPT-1.3B-std yields competitive results against OPT-1.3B and Pythia-1.4B.
% Speed-up results and some interesting phenomena.

\paragraph{Massive Multitask Language Understanding.}
We also evaluate our models on the massive multitask language understanding benchmark (MMLU) \citep{hendrycks2020measuring} which consists of multiple-choice questions covering diverse domains of knowledge, such as humanities, STEM, and social sciences. At evaluation time, we use the examples provided by the benchmark, and the results of our models on the MMLU benchmark are reported in Table \ref{tab2}. 

On this benchmark, RecycleGPT-1.3B outperforms OPT-1.3B and Pythia-1.4B and is Slightly lower than GPT-Neo-2.7B due to parameter size. 
Compared with the zero-shot setting, our RecycleGPT can achieve better results on the few-shot setting. A potential explanation is that our method is more applicable to situations with more examples or demonstrations due to the model architecture and decoding strategy we designed. Or perhaps our approach can better model certain types of context.
This phenomenon also guides us on how to better utilize and improve our methods in the future. 
The detailed performance results on the 57 tasks of MMLU can be found in Table \ref{MMLU} in the appendix.

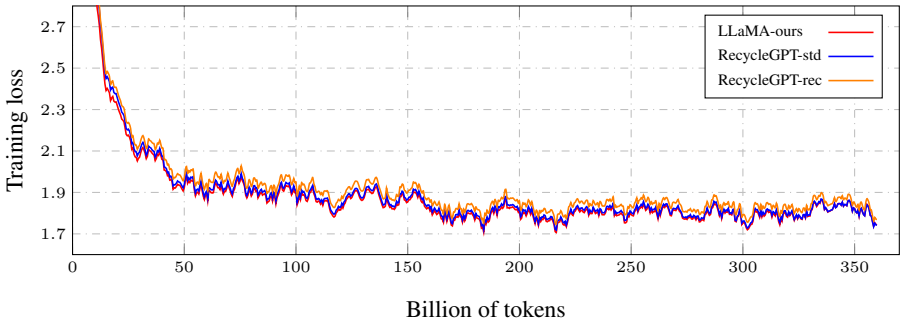
\begin{figure}[!t]
    \centering
    \begin{tikzpicture}
      \scriptsize{
      \begin{axis}[
      width=.9\textwidth, height=.35\textwidth ,
      xlabel={\small{Billion of tokens}},
      ylabel={\small{Training loss}},
      xmin=0, xmax=370,
      ymin=1.6, ymax=2.8,
      xtick={0,50,100,150,200,250,300,350},
      ytick={1.7,1.9,2.1,2.3,2.5,2.7},
      yticklabels={$1.7$,$1.9$,$2.1$,$2.3$,$2.5$,$2.7$},
      ymajorgrids=true,
      xmajorgrids=true,
      grid style=dashdotted,
      legend cell align=left,
      scaled ticks=false,
      xlabel style={align=center,font=\scriptsize},
      ylabel style={font=\scriptsize,yshift=-2em},
      y tick style={opacity=0},
      x tick label style={font=\tiny},
      y tick label style={font=\tiny},
      legend style={yshift=-0.2em,xshift=0em,legend cell align=left,legend plot pos=right},
      ]
      \addplot [sharp plot,red,mark size=1pt,thick,line width=0.6pt,mark size=0.2pt] table [x=Step,y=Value,col sep=comma] {base_smooth.csv};
      \addplot [sharp plot,blue,mark size=1pt,thick,line width=0.6pt,mark size=0.2pt] table [x=Step,y=Value,col sep=comma] {std_smooth.csv};
      \addplot [sharp plot,orange,mark size=1pt,thick,line width=0.6pt,smooth] table [x=Step,y=Value,col sep=comma] {rec_smooth.csv};
      \legend{\tiny{LLaMA-ours},\tiny{RecycleGPT-std},\tiny{RecycleGPT-rec}},
      \end{axis}
      }
      % \pgfmathprintnumber[fixed zerofill,precision=5]{12345}
      \end{tikzpicture}
    \caption{Training loss over train tokens.}
     \label{fig:loss}
\end{figure}

Figure \ref{fig:loss} plots the training loss of the baseline, RecycleGPT-std, and RecycleGPT-rec. We can see that the training loss of baseline and RecycleGPT-std are almost identical which proves that our approach does not impact the performance of the original language model. At the same time, we also see that the curves of RecycleGPT-rec and baseline are very close. It demonstrates the effectiveness of our method.
We report the speed (ms/token) of our RecycleGPT in table \ref{tab_speed}.
RecycleGPT achieves a 1.4x speedup over the baseline model with KV cache and a 1.34x speedup without KV cache.
The experiments in the current work were conducted on a 1.3B model due to computational constraints. In future work, we will experiment on larger models, such as 7B and 13B.

\begin{table*}[!ht]
\centering
\small
\begin{spacing}{1.1}
\setlength{\tabcolsep}{1.3mm}{
\begin{tabular}{lccccccccc}
\toprule
{\bf Model} & {} & {\bf Humanities} & {\bf STEM} & {\bf Social Sciences} & {\bf Other} & {\bf Average} \\
\midrule
OPT & 1.3B & 22.8 & 25.7 & 23.3 & 26.5 & 24.6  \\
Pythia & 1.4B & 26.6 & 25.6 & 24.3 & 26.6 & 25.8  \\
GPT-Neo & 2.7B & 25.3 & 25.6 & 27.5 & 27.4 & 26.4  \\
\midrule
LLaMA-ours & 1.3B & 27.8 & 26.1 & 23.5 & 23.7 & 25.4  \\
RecycleGPT-std & 1.3B & 26.5 & 28.2 & 24.0 & 25.0 & 26.2  \\
RecycleGPT-rec & 1.5B & 26.3 & 28.0 & 24.0 & 24.8 & 26.0  \\
\bottomrule 
\end{tabular}}
\end{spacing}
\caption{
\label{tab2}
{\bf Five-shot performance on Massive Multitask Language Understanding (MMLU).}
}
\end{table*}

\begin{table*}[!ht]
\centering
\small
\begin{spacing}{1.1}
\setlength{\tabcolsep}{1.3mm}{
\begin{tabular}{lccccccc}
\toprule
\multirow{2}{*}{\bf Model} & \multicolumn{5}{c}{ms/token} & \multirow{2}{*}{\bf Avg} & 
 \multirow{2}{*}{\bf Avg Speed Up} \\
\cmidrule(r){2-6}
& 64 & 128 & 256 & 512 & 1024 \\
\midrule
\multicolumn{7}{l}{KV cache} \\
RecycleGPT-std & 18.4 & 19.2 & 18.7 & 18.5 & 18.6 & 18.7 & 1X \\
RecycleGPT-rec & 13.8 & 13.1 & 13.4 & 13.0 & 13.7 & 13.4  & 1.40X \\
\midrule
\multicolumn{7}{l}{w/o KV cache} \\
RecycleGPT-std & 20.8 & 24.1 & 33.0 & 55.3 & 103.7 & 47.4 & 1X \\
RecycleGPT-rec & 14.8 & 16.6 & 24.4 & 41.4 & 80.4 & 35.5 & 1.34X \\
\bottomrule
\end{tabular}}
\end{spacing}
\caption{
\label{tab_speed}
{\bf Decoding speed of RecycleGPT-std and RecycleGPT-rec at different sequence lengths}.
}
\end{table*}

\section{Related Work}
The scale of auto-regressive language models grows from 117M \citep{radford2018improving} parameters to over 500B parameters \citep{smith2022using} and various approaches are explored to improve the inference efficiency. 
Large amounts of model computations and memory movements are the key factors of slower inference \citep{pope2023efficiently}.
To make model size smaller, several works are proposed distillation \citep{hinton2015distilling,sanh2019distilbert}, pruning \citep{li2020efficient,brix2020successfully,zhou2021learning}, sharing weights \citep{xiao2019sharing} or quantization to int8 or even int4 \citep{dettmers2022llm,shen2020q,zafrir2019q8bert,zafrir2019q8bert}.
Adaptive computations \citep{sukhbaatar2019adaptive,schwartz2020right} try to reduce the amount of computation for easier inference steps.
\citet{sukhbaatar2019adaptive,kitaev2020reformer,zeng2021recurrent,roy2021efficient,choromanski2020rethinking} propose efficient attention layers to overcome the computational bottlenecks that time and memory scales quadratic in the sequence length.
Based on the memory complexity of self-attention layers, \citet{dao2022flashattention,shazeer2019fast} propose new attention algorithms to reduce the number of memory reads/writes
between (HBM) and GPU on-chip SRAM.

Apart from improving the model architecture for faster decoding, sampling strategies, and partitioning strategies can also achieve low-latency inference \citep{stern2018blockwise,ge2022lossless}. Speculative sampling methods employ multiple small efficient models to generate draft tokens and thus, run fewer forward calls of large model \citep{chen2023accelerating,leviathan2023fast,miao2023specinfer}.
For larger models that fit on different accelerator chips,
practical partitioning approaches are proposed for balance workloads \citep{pope2023efficiently}.
This work also tries to minimize the number of forward calls of language models.
Compared to previous methods that reduce the number of model invocations, such as speculative decoding \citep{chen2023accelerating,leviathan2023fast}, our method is fine-grained while also being orthogonal to their methods, allowing further acceleration on top of them.

\section{Conclusion}
 % we design a new LM, ... 2 times ...
In this work, we propose RecycleGPT, a new architecture with low-inference latency. By predicting multiple tokens with the recyclable module at once, RecycleGPT can achieve up to 1.4x speedup with no performance loss. The proposed approach is model-agnostic and complementary to previous acceleration techniques. In the future, we will explore more decoding strategies by combining the recyclable module and the original model in various ways.

\bibliography{iclr2023_conference}
\bibliographystyle{iclr2023_conference}

\appendix
\section{Appendix}
%You may include other additional sections here.
\begin{table*}[!h]
\centering
\small
\begin{spacing}{1.2}
\begin{tabular}{lccc}
\toprule
{\bf Pre-training Hyperparameters} & {\bf 1.3B} \\
\midrule
Number of layers & 24  \\
Hidden Size & 2048  \\
FFN inner hidden size & 5504  \\
Attention heads & 32  \\
Attention head size & 64  \\
Embedding Size & 2048  \\
Warmup steps & 1.5k  \\
Learning Rate & 2e-4  \\
Adam $\epsilon$ & 1e-8  \\
Adam $\beta_1$ & 0.9  \\
Adam $\beta_2$ & 0.95  \\
Attention Dropout & 0.0  \\
Dropout & 0.0  \\
Weight Decay & 0.01  \\
Max Sequence Length & 2048  \\
Batch Size & 1280 \\
Train Steps & 140k \\
RmsNorm eps & 1e-06 \\
\bottomrule 
\end{tabular}
\end{spacing}
\caption{
\label{tab_settings_pretraining}
{\bf The pre-training hyperparameters}.
}
\end{table*}

\begin{table*}[!ht]
\centering
\tiny
\begin{spacing}{1.2}
\setlength{\tabcolsep}{1.2mm}{
\begin{tabular}{lccccccc}
\toprule
{} & {} & {\bf OPT-1.3B} & {\bf Pythia-1.4B} & {\bf GPT-Neo} & {\bf LLaMA-ours} & {\bf RecycleGPT-std} & {\bf RecycleGPT-rec}  \\
\midrule
Abstract Algebra & STEM & 0.29 & 0.25 & 0.25 & 0.34 & 0.28 & 0.28 \\
Anatomy & STEM & 0.2815 & 0.3185 & 0.2074 & 0.3556 & 0.3259 & 0.3259 \\
Astronomy & STEM & 0.2039 & 0.25 & 0.1908 & 0.3421 & 0.2632 & 0.25 \\
Business Ethics & Other & 0.22 & 0.23 & 0.29 & 0.38 & 0.25 & 0.23 \\
Clinical Knowledge & Other & 0.2453 & 0.2528 & 0.2642 & 0.3245 & 0.2264 & 0.2189 \\
College Biology & STEM & 0.2569 & 0.3056 & 0.2569 & 0.3611 & 0.2431 & 0.2569 \\
College Chemistry & STEM & 0.17 & 0.21 & 0.22 & 0.39 & 0.27 & 0.24 \\
College Computer Science & STEM & 0.26 & 0.27 & 0.25 & 0.28 & 0.32 & 0.37 \\
College Mathematics & STEM & 0.23 & 0.24 & 0.28 & 0.3 & 0.3 & 0.28 \\
College Medicine & Other & 0.2543 & 0.2543 & 0.2428 & 0.3121 & 0.2312 & 0.1908 \\
College Physics & STEM & 0.2843 & 0.2451 & 0.2255 & 0.2353 & 0.2549 & 0.2843 \\
Computer Security & STEM & 0.18 & 0.22 & 0.28 & 0.46 & 0.3 & 0.3 \\
Conceptual Physics & STEM & 0.2213 & 0.3106 & 0.2596 & 0.366 & 0.2085 & 0.2511 \\
Econometrics & Social Science & 0.2719 & 0.2281 & 0.2632 & 0.2632 & 0.2456 & 0.2281 \\
Electrical Engineering & STEM & 0.3034 & 0.2621 & 0.2552 & 0.2483 & 0.2414 & 0.2483 \\
Elementary Mathematics & STEM & 0.254 & 0.2672 & 0.2937 & 0.2619 & 0.2487 & 0.2646 \\
Formal Logic & Humanities & 0.1349 & 0.127 & 0.1825 & 0.2063 & 0.1508 & 0.1508 \\
Global Facts & Other & 0.35 & 0.36 & 0.2 & 0.31 & 0.32 & 0.31 \\
High School Biology & STEM & 0.2194 & 0.2548 & 0.2484 & 0.3613 & 0.2581 & 0.2613 \\
High School Chemistry & STEM & 0.2709 & 0.2512 & 0.2414 & 0.2808 & 0.3005 & 0.2562 \\
High School Computer Science & STEM & 0.34 & 0.27 & 0.35 & 0.31 & 0.3 & 0.32 \\
High School European History & Humanities & 0.2303 & 0.2545 & 0.2303 & 0.4485 & 0.2485 & 0.2909 \\
High School Geography & Social Science & 0.2525 & 0.2424 & 0.3283 & 0.3535 & 0.2576 & 0.2576 \\
High School Government And Politics & Social Science & 0.2694 & 0.1917 & 0.2591 & 0.4508 & 0.2642 & 0.2539 \\
High School Macroeconomics & Social Science & 0.2538 & 0.2128 & 0.3487 & 0.3436 & 0.2103 & 0.2103 \\
High School Mathematics & STEM & 0.237 & 0.2444 & 0.2481 & 0.2407 & 0.2556 & 0.2556 \\
High School Microeconomics & Social Science & 0.1975 & 0.2269 & 0.2395 & 0.3319 & 0.2185 & 0.2185 \\
High School Physics & STEM & 0.2848 & 0.245 & 0.2384 & 0.2781 & 0.2848 & 0.2914 \\
High School Psychology & Social Science & 0.244 & 0.2569 & 0.3064 & 0.4624 & 0.2294 & 0.2477 \\
High School Statistics & STEM & 0.2639 & 0.2454 & 0.4074 & 0.3519 & 0.412 & 0.3333 \\
High School Us History & Humanities & 0.1814 & 0.2843 & 0.201 & 0.3578 & 0.2549 & 0.2451 \\
High School World History & Humanities & 0.2405 & 0.2574 & 0.2194 & 0.4388 & 0.2785 & 0.2489 \\
Humanities Aging & Other & 0.2646 & 0.3274 & 0.1839 & 0.417 & 0.287 & 0.278 \\
Humanities Sexuality & Social Science & 0.2366 & 0.2519 & 0.2748 & 0.3817 & 0.2748 & 0.2214 \\
International Law & Humanities & 0.2727 & 0.3554 & 0.2314 & 0.5537 & 0.3802 & 0.3719 \\
Jurisprudence & Humanities & 0.2222 & 0.25 & 0.2963 & 0.4352 & 0.2685 & 0.2222 \\
Logical Fallacies & Humanities & 0.2515 & 0.3006 & 0.2577 & 0.4233 & 0.3129 & 0.3067 \\
Machine Learning & STEM & 0.2321 & 0.2054 & 0.1696 & 0.2411 & 0.2946 & 0.25 \\
Management & Other & 0.1553 & 0.2524 & 0.2718 & 0.3592 & 0.2039 & 0.165 \\
Marketing & Other & 0.235 & 0.2607 & 0.265 & 0.4615 & 0.265 & 0.2863 \\
Medical Genetics & Other & 0.27 & 0.26 & 0.29 & 0.4 & 0.26 & 0.23 \\
Miscellaneous & Other & 0.2746 & 0.2746 & 0.2363 & 0.4317 & 0.2682 & 0.2695 \\
Moral Disputes & Humanities & 0.2341 & 0.2775 & 0.2457 & 0.3815 & 0.2601 & 0.2341 \\
Moral Scenarios & Humanities & 0.2447 & 0.248 & 0.2704 & 0.2425 & 0.2458 & 0.2514 \\
Nutrition & Other & 0.2745 & 0.2582 & 0.317 & 0.3922 & 0.2614 & 0.2843 \\
Philosophy & Humanities & 0.1961 & 0.2926 & 0.3248 & 0.4116 & 0.299 & 0.299 \\
Prehistory & Humanities & 0.2747 & 0.2562 & 0.3056 & 0.3488 & 0.2407 & 0.2778 \\
Professional Accounting & Other & 0.2553 & 0.266 & 0.2553 & 0.2766 & 0.2553 & 0.2695 \\
Professional Law & Humanities & 0.2269 & 0.2627 & 0.2477 & 0.2934 & 0.2451 & 0.2471 \\
Professional Medicine & Other & 0.375 & 0.1838 & 0.4301 & 0.4449 & 0.1765 & 0.2426 \\
Professional Psychology & Social Science & 0.268 & 0.2729 & 0.2696 & 0.3578 & 0.2516 & 0.268 \\
Public Relations & Social Science & 0.1727 & 0.3091 & 0.1818 & 0.3727 & 0.2 & 0.2364 \\
Security Studies & Social Science & 0.2041 & 0.2041 & 0.2939 & 0.3388 & 0.2327 & 0.2286 \\
Sociology & Social Science & 0.2338 & 0.2637 & 0.2289 & 0.4726 & 0.2537 & 0.2637 \\
Us Foreign Policy & Social Science & 0.2 & 0.26 & 0.31 & 0.45 & 0.24 & 0.24 \\
Virology & Other & 0.2771 & 0.2771 & 0.3193 & 0.3193 & 0.247 & 0.247 \\
World Religions & Humanities & 0.2573 & 0.2924 & 0.2807 & 0.4795 & 0.2573 & 0.2749 \\
\bottomrule 
\end{tabular}}
\end{spacing}
\caption{
\label{MMLU}
{\bf Detailed five-shot results per domain on the MMLU test sets.}
}
\end{table*}

\end{document}